\documentclass[10pt,twocolumn,letterpaper]{article}

\usepackage{cvpr}              % To produce the CAMERA-READY version
\definecolor{cvprblue}{rgb}{0.21,0.49,0.74}
\usepackage[pagebackref,breaklinks,colorlinks,allcolors=cvprblue]{hyperref}
\usepackage{hyperref}
\usepackage[ruled,vlined,linesnumbered]{algorithm2e}
\usepackage{multirow}
\usepackage{amsmath}
\usepackage[accsupp]{axessibility}  % Improves PDF readability for those with disabilities.

\definecolor{myorange}{RGB}{255, 128, 0}
\definecolor{mygreen}{RGB}{0, 175, 0}
\definecolor{myblue}{RGB}{0, 102, 204}
\definecolor{lightblue}{RGB}{232, 244, 248}
\definecolor{myred}{RGB}{220, 60, 60}
\definecolor{mypurple}{RGB}{128, 0, 128}

\definecolor{greenyellow}{RGB}{197,228,10}

\def\paperID{2278} % *** Enter the Paper ID here
\def\confName{CVPR}
\def\confYear{2025}

\title{VasTSD: Learning 3D Vascular Tree-state Space Diffusion Model for \\ Angiography Synthesis}
\author{
Zhifeng Wang\footnotemark[1] \quad
Renjiao Yi\footnotemark[1] \quad
Xin Wen \quad 
Chenyang Zhu\footnotemark[2] \quad
Kai Xu\footnotemark[2] \\[2.5mm]
\textsuperscript{}National University of Defense Technology\\ 
}

\begin{document}
\maketitle

% %%%%%%%%%%%%%%%%%%%%% foot note for authors
% \renewcommand{\thefootnote}{}  % 让脚注不带编号
% \footnote{$^{\ast}$Co-first authors. $^{\dagger}$Corresponding authors.}
% % \footnote{$^{\dagger}$Corresponding author.}
\footnote{$^{\ast}$Co-first authors. $^{\dagger}$Corresponding authors. Corresponding authors: \texttt{zhuchenyang07@nudt.edu.cn}, \texttt{kevin.kai.xu@gmail.com}.}

\begin{abstract}
Angiography imaging is a medical imaging technique that enhances the visibility of blood vessels within the body by using contrast agents. Angiographic images can effectively assist in the diagnosis of vascular diseases. However, contrast agents may bring extra radiation exposure which is harmful to patients with health risks. To mitigate these concerns, in this paper, we aim to automatically generate angiography from non-angiographic inputs, by leveraging and enhancing the inherent physical properties of vascular structures. Previous methods relying on 2D slice-based angiography synthesis struggle with maintaining continuity in 3D vascular structures and exhibit limited effectiveness across different imaging modalities. We propose VasTSD, a 3D vascular tree-state space diffusion model to synthesize angiography from 3D non-angiographic volumes, with a novel state space serialization approach that dynamically constructs vascular tree topologies, integrating these with a diffusion-based generative model to ensure the generation of anatomically continuous vasculature in 3D volumes. A pre-trained vision embedder is employed to construct vascular state space representations, enabling consistent modeling of vascular structures across multiple modalities. Extensive experiments on various angiographic datasets demonstrate the superiority of VasTSD over prior works, achieving enhanced continuity of blood vessels in synthesized angiographic synthesis for multiple modalities and anatomical regions.
\end{abstract}    
\section{Introduction}
\label{sec:intro}
Vascular diseases are widely recognized as major contributors to human mortality worldwide~\cite{update2017heart,wang2024cardiovascular}. In clinical practice, angiography serves as the gold standard for diagnosing vascular conditions and diseases, such as coronary artery disease, atherosclerosis, and aneurysms. In scientific research, angiography provides high-resolution imaging of vascular anatomy, aiding researchers in the detailed exploration of vascular physiology, hemodynamic features, and more, thereby offering crucial data for fundamental research. 
Traditional angiographic methods, such as X-ray Angiogram (XRA), Computed Tomography Angiography (CTA), and Magnetic Resonance Angiography (MRA), have demonstrated their effectiveness, while they require the use of nephrotoxic iodinated contrast agents and expose patients to ionizing radiation, leading to potential carcinogenic risks~\cite{weisbord2020contrast,zanzonico2006radiation}.%, ultimately posing additional health risks for patients. 
Thus, angiography synthesis from non-angiography data becomes a critical direction.

\begin{figure}
  \centering
  \includegraphics[width=\linewidth]{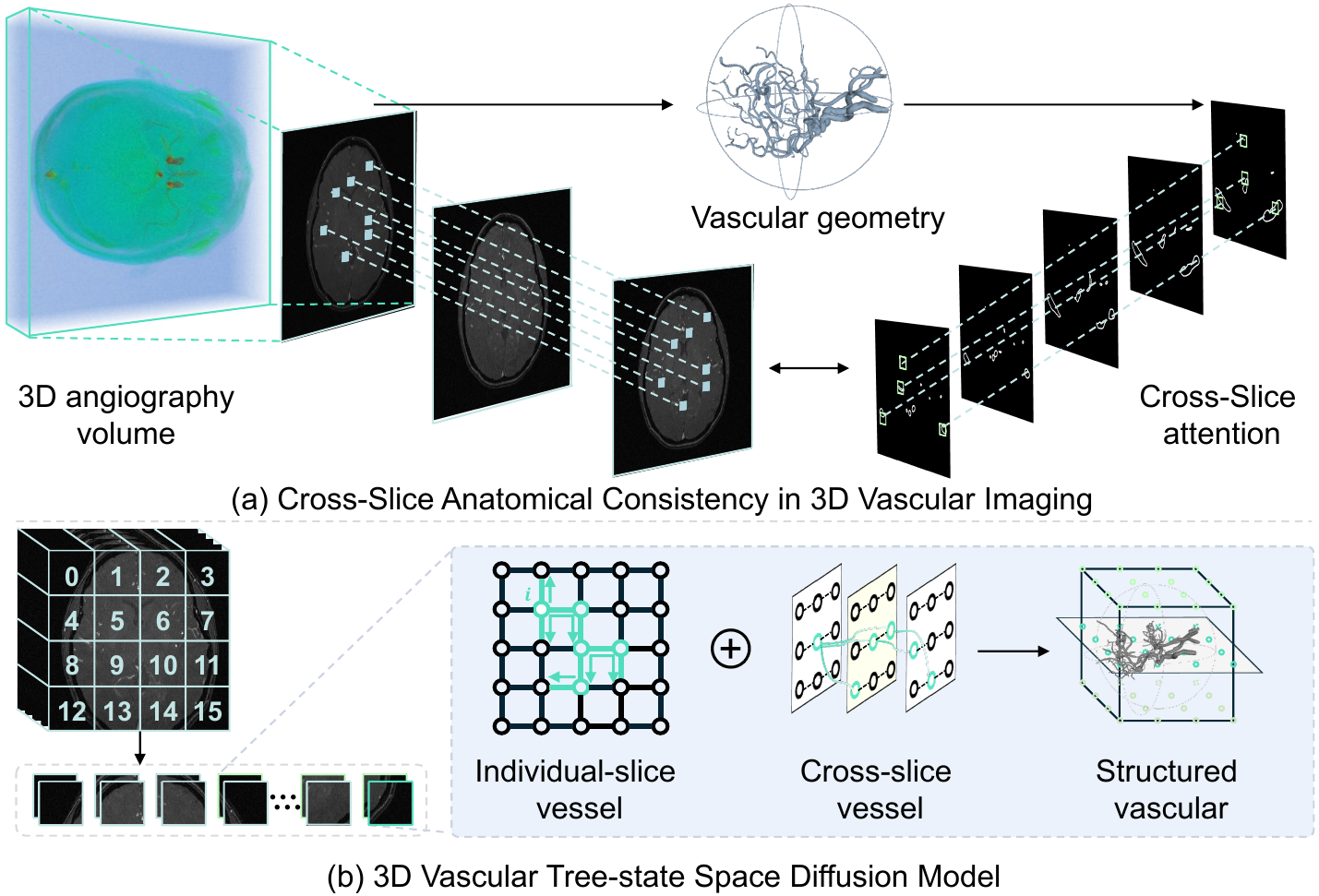} 
  \caption{(a) In 3D medical volumes, vascular have relatively consistent contextual semantics and structural connectivity. (b) The proposed tree-state space model for simulating 3D vasculature.}
  \label{fig:introduction}
  \vspace{-0.3cm}
\end{figure}

With the advancement of generative deep learning technologies, the synthesized medical images become more realistic and diverse. Existing medical image generation methods are broadly categorized into several types based on backbone models, including Variational Autoencoders (VAEs)~\cite{kingma2013auto,pesteie2019adaptive}, Generative Adversarial Networks (GANs)~\cite{goodfellow2020generative,yi2019generative}, and Diffusion Models~\cite{ho2020denoising,kazerouni2023diffusion}. Recent approaches~\cite{lyu2023generative,koch2024cross,wang2024cardiovascular,zhou2020hi} also explored cross-modal image transformation. For instance, MedGAN~\cite{armanious2020medgan} enables multi-modal conversion by learning end-to-end image transformation. SynDiff~\cite{ozbey2023unsupervised} employs a conditional diffusion process to capture the correlations in image distributions, allowing for unsupervised multi-modal synthesis. These methods could all potentially be used for angiography synthesis. However, we observe that these image-based generation approaches synthesize incontinous blood vessels in 3D volume. Moreover, each method only works for transformation for one pair of modalities. 

We believe that due to variations in imaging devices and principles, the data distributions of different types of medical images exhibit significant disparities within 3D space. Nevertheless, within the same modality, vascular data distributions demonstrate geometric continuity. Thus, integrating medical data from different modalities into a unified state space may be feasible for different multi-modal transformation tasks, and by incorporating the geometric representation of blood vessels and high-level vascular semantic information, continuity of vessels in angiography synthesis can be achieved. 

% To this end, this paper 
As shown in Fig.~\ref{fig:introduction}(a), we observe that in 3D vascular images, the vascular structures (semantic regions) exhibit consistent positioning across slices and share relatively consistent anatomical characteristics (shape). Therefore, the geometric state space consistency across 3D vascular sequences offers a promising direction. 
This paper proposes a cross-slice state space vision embedding approach aimed at unifying vascular scenes by encoding different sequences of vascular images into the same state space. 
This unified visual encoding is adaptable for various modalities of vascular data. 
As illustrated in Fig.~\ref{fig:introduction}(b), the geometric structure of blood vessels exists not only within individual slices but also across slices. 
Masks are guided by a vision embedder to enhance the cross-slice consistency. A tree topology is introduced based on the spatial relationships of the vessels and their input features within the same slice.
This approach is implemented by linear complexity dynamic programming, effectively breaking the traditional sequential constraints.
A tree-state space model is introduced to facilitate long-range interactions in 3D angiography, enhancing the geometric integrity of synthesized vessels.

As illustrated in Fig.~\ref{fig:framework}, this paper introduces a 3D \textbf{Vas}cular \textbf{T}ree-state \textbf{S}pace \textbf{D}iffusion model (VasTSD) for synthesizing vascular structures in angiography. In particular, we first utilize a pre-trained vision embedder to process slices from different slices, extracting token weights that are subsequently integrated into a diffusion model for angiography synthesis. This framework supports multi-modal angiographic synthesis tasks, offering a novel solution for clinical diagnostics and fundamental research in angiography. 
To quantify the effectiveness, comprehensive comparisons are conducted with several prior approaches of medical modality conversion. Experiments demonstrate that we achieved state-of-the-art results on commonly used angiography datasets. Furthermore, we provide a clear 3D visualizations of synthesized angiographic data, showing the continuity of blood vessels within resulting 3D volumes. The contributions are summarized as follows:
\begin{itemize}
    \item VasTSD is the first framework based on state space model for synthesizing angiographies, achieving high structural fidelity from non-angiographic data across multiple modalities and anatomical regions.
    \item A dynamic spanning tree of angiography state space features is introduced based on dynamic programming to achieve continuous and semantic long-range interaction of vascular geometry. 
    \item This paper presents a pre-trained vision embedder mechanism that ensures the geometric consistency of blood vessels and continuous structure representation with directional semantic propagation in angiography.
\end{itemize}

\section{Related Works}
\subsection{Deep Learning for Angiography Synthesis}
Angiography synthesis has gained attention in computer-aided diagnosis, driven by recent advancements in deep learning. Current research focuses on two main approaches to achieve high-quality synthesis.
The first approach centres on modality conversion via generative methods within a single modality~\cite{kshirsagar2024generative,koch2024cross,xia2023virtual}. For instance, Lyu et al.~\cite{lyu2023generative} proposed a generative adversarial method to synthesize CTA from CT. Similarly, Kamran et al.~\cite{kamran2021attention2angiogan} introduced Attention2AngioGAN, which enables the synthesis of fluorescein angiography from retinal fundus images. 
The second approach focuses on generating high-quality non-angiographic angiograms. For example, Javier et al.~\cite{montalt2021reducing} achieved high-quality cardiovascular angiography synthesis using MRI with a simple residual U-Net model. Meanwhile, Lyu et al.~\cite{lyu2023head} introduced a denoising diffusion probabilistic model to enhance contrast agents in Head-Neck Dual-energy CT, achieving high-quality contrast-enhanced vasculature with only 12.5\% of the normal dosage. These methods are effective for specific single-modality synthesis, and vascular geometry is not considered. 

\begin{figure*}
  \centering
  \includegraphics[width=\textwidth]{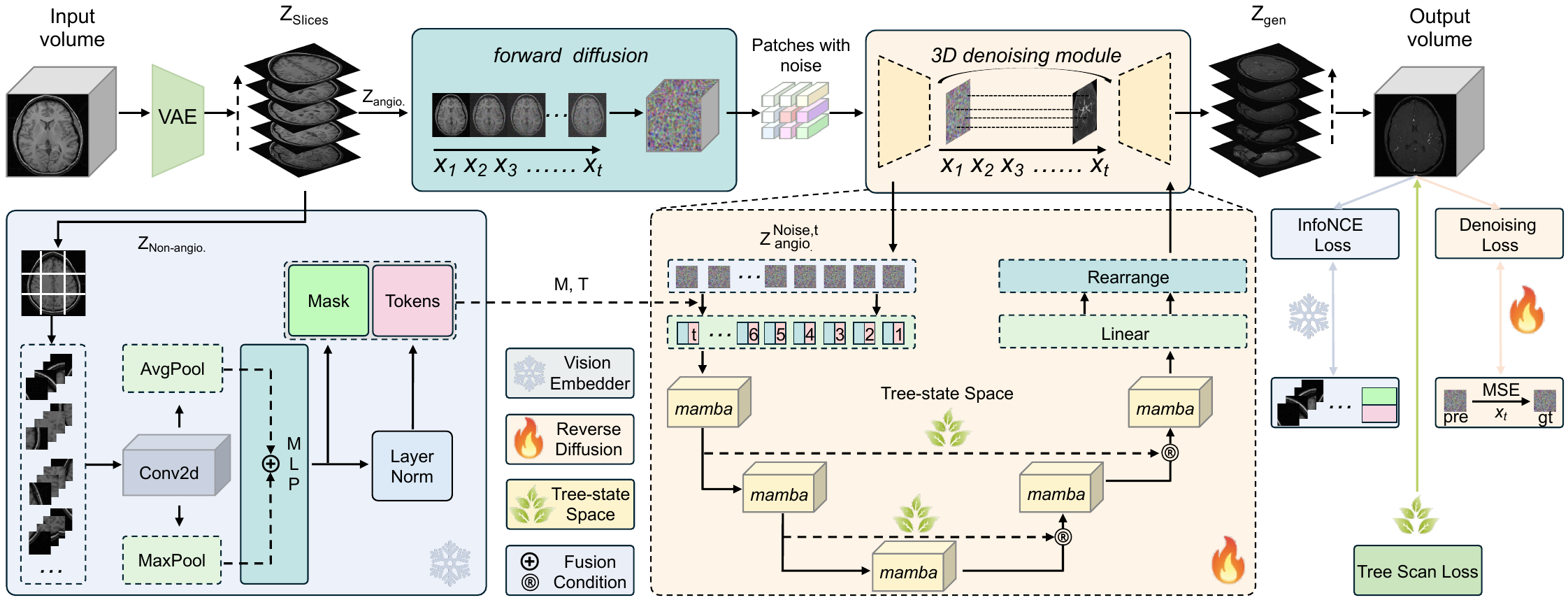} 
  \caption{\textbf{The overall framework of VasTSD.} VasTSD contains a pre-trained vision embedder and a 3D vascular tree-state space diffusion module. The vision embedder encodes 3D medical data and generates embeddings for the diffusion process. The 3D vascular state space diffusion module consists of a forward diffusion and a 3D denoising process based on tree-state space.}
  \label{fig:framework}
  \vspace{-0.3cm}
\end{figure*}
\vspace{-0.1cm}
\subsection{Diffusion Models for Medical Image Synthesis}
Diffusion models have gained significant attention in medical imaging~\cite{kazerouni2023diffusion,deshpande2024assessing,ozbey2023unsupervised} for their ability to generate high-quality images and model complex data distributions, with successful applications in tasks like image reconstruction~\cite{song2024diffusionblend}, denoising~\cite{gao2023corediff}, and segmentation~\cite{liu2024diffrect}. Recent studies have demonstrated their potential in generating synthetic medical images, particularly for modalities like MRI, CT, and X-ray, where high-quality data is often limited. For instance, Ho et al.~\cite{chung2022score} introduced a score-based generative model demonstrating remarkable results in generating realistic MRI scans. DiffuseReg~\cite{zhuo2024diffusereg} and DRDM~\cite{zheng2024deformation} generate medical deformation fields for medical image synthesis but are not designed for angiography-related tasks. SynDiff~\cite{ozbey2023unsupervised} proposed an unsupervised diffusion-based method for high-fidelity medical image synthesis. Fast-DDPM~\cite{jiang2024fast} focuses on improving the speed of image denoising and image-to-image translation tasks. DiffMa~\cite{wang2024soft} aims to integrate mamba to achieve modality conversion from CT to MRI. While promising in medical image synthesis, diffusion models are sensitive to noise due to their generation process, which can cause information loss and feature blurring. In tasks like angiographic synthesis, incorporating physical characteristics to guide the process is crucial to mitigate these issues.

\subsection{State Space Models (SSMs)}
State Space Models (SSMs)~\cite{liu2024vmamba,zhu2024vision,huang2024localmamba,he2024multi,hu2024zigma} are a novel deep learning approach for sequence-to-sequence tasks, particularly effective with long sequences. Structured State Space Models were introduced to address the challenge of long-range dependencies by utilizing a structured state space framework, sparking a wave of subsequent research in the area. Recently, several SSMs-based approaches have been developed for medical applications, including tasks such as medical image segmentation~\cite{wang2024mamba,tsai2024uu,ruan2024vm}, image denoising~\cite{ozturk2024denomamba}, and classification~\cite{yue2024medmamba,gong2024nnmamba}, achieving notable success. These methods typically employ fixed scanning mechanisms to systematically transfer information between time steps, effectively capturing dependencies within sequences. However, they struggle to capture structures and spatial relationships inherent in 3D vascular data. Thus, we propose a dynamic tree-state space model to enhance long-range modeling of vascular geometry.

\section{Method}\label{sec:overview}
%-------------------------------------------------------------------------
%\subsection{Overview}
Fig.~\ref{fig:framework} presents the overall framework of VasTSD. 
It builds upon the foundational principles of diffusion models, overcoming the heterogeneity between different modalities and anatomical regions while addressing key limitations in 3D structured generation within the medical field. 
The vision embedder is presented in Sec.~\ref{sec:embedder}. 
Given 3D medical angiography and non-angiography volumes as input, we first utilize a Variational Autoencoder (VAE) to transform the data into a 3D latent space representation along the cross-sectional slices, denoted as $Z_{Slices}$. Next, we perform a vision embedder on the non-angiographic data to generate the conditional embedding representations of the mask $M$ and tokens $T$, which are then used as inputs to the diffusion process. The diffusion module employed for angiography synthesis is described in Sec.~\ref{sec:diffusion_related}. At each forward diffusion step, we apply noise to the angiography and perform reverse denoising by incorporating a tree-state space to synthesize new angiograms. Here, a novel vascular dynamic tree topology scanning method is proposed, as shown in Fig.~\ref{fig:mamba}, which integrates the 3D physical properties of the vasculature to enable high-quality 3D vascular synthesis, as detailed in Sec.~\ref{sec:ssm}. At last, the 3D angiographic volumes are reconstructed by VAEs.
%-------------------------------------------------------------------------
\subsection{Pre-trained Vision Embedder}\label{sec:embedder}
The inherent vascular location features in angiography provide prior knowledge for synthesizing continuous angiography from non-angiography. 
We leverage this by designing a pre-trained vision embedder that effectively captures vascular structures. This embedder extracts high-quality representations of vascular regions from medical slices, forming the foundation for the reverse diffusion process. 
Specifically, the vision embedder converts non-angiographic 2D vessel slices into latent feature representations, reducing the data dimensionality and complexity. It employs average- and max-pooling to capture global and local features. These pooled features are then passed through a multi-layer perception, which maps them into a specific embedding space.

Additionally, a contrastive learning framework is incorporated by introducing an Exponential Moving Average mechanism to smooth the vascular sequences. The InfoNCE loss is employed to compute both the similarity and dissimilarity between samples. A normalized feature similarity matrix is then computed from the processed features, and cross-entropy loss is applied to guide the model in learning more discriminative feature representations. Thus, the loss function $\mathcal{L}_{\text{InfoNCE}}$ is formulated as:
\begin{equation}\label{eq:infonce_loss}
\mathcal{L}_{\text{InfoNCE}} = -\log \frac{\exp(\text{sim}(z_i, z_j) / \tau)}{\sum_{k=1}^{N} \exp(\text{sim}(z_i, z_k) / \tau)},
\end{equation}
where $z_i$ represents the latent representation $Z_{Non-angio.}$ by VAE from non-angiography. $z_j$ denotes the tokenized representation of the same input image obtained through the MLP, and sim$(z_i, z_j)$ represents the similarity measure between $z_i$ and $z_j$, 
and $z_k$ represents the latent representations of other images in the batch, serving as negative samples in contrast to the positive pair $z_i$ and $z_j$.

The visual embedding model was trained on paired angiographic and non-angiographic data of the same modality. From each non-angiographic slice, vascular positional and structural information was extracted to generate embedded representations of the mask ($M$) and labels ($T$) for all slice elements. These embeddings provide spatial constraints that guide the diffusion process, ensuring the synthesized angiography aligns with the vascular structure and enhancing ability of the model to capture detailed features. More details are in supplementary materials.

%-------------------------------------------------------------------------
\subsection{Diffusion Module for Angiography Synthesis}\label{sec:diffusion_related}
A 3d tree-state space diffusion model is introduced to synthesize angiographic data from features of the vision embedder. 
This module consists of two phases: forward diffusion and 3D denoising. In the reverse denoising phase, we propose a tree-state space model to remove noise, generating high-quality angiography iteratively. 

\noindent\textbf{Forward diffusion.} The latent representation $z_{\text{angio.}}$ of the angiography is generated by the Variational Autoencoder (VAE). This representation, along with the time step $t$, serves as input to the forward diffusion process. In this process, noise is progressively introduced in successive steps.
These processes can be parameterized as:
\begin{equation}\label{eq:forward_diffusion_conditioned}
z^t_{\text{angio.}} = \sqrt{\alpha_t} \cdot z_{\text{angio.}} + \sqrt{1 - \alpha_t} \epsilon,
\end{equation}
where $t$ represents the time step in the diffusion process, and $\alpha_t$ is a scaling factor that controls the amount of original signal versus noise at each step, $\epsilon$ denotes the noise sampled from a gaussian distribution with the same dimensions as the latent representation $z_{\text{angio.}}$.

\noindent\textbf{3D denoising.} The angiography noise slices generated by the forward diffusion process are divided into blocks $z^{Noise,t}_{\text{angio.}}$, which serve as input for the reverse denoising process. Meanwhile, non-angiography features from the pre-trained vision embedder provide crucial embedding representations for reverse diffusion. During the reverse denoising process, noise is gradually removed, and angiography is progressively reconstructed. The vascular tree-state space model enables hierarchical conditional feature aggregation, ensuring the seamless merging of embedded representations $M$ and $T$ at each restoration step. Finally, after linear transformation, rearrangement, and optimization via conditional fusion, the final angiography slice $z_{\text{gen}}$ is generated.

In the reverse diffusion process, the Mean Squared Error (MSE) loss minimizes the difference between the generated image $x_t$ at each timestep and the target noise-free image $x_0$, guiding the model to eliminate noise progressively.
The Denoising loss is defined as:
\begin{equation}\label{eq:MSE_loss}
\mathcal{L}_{\text{diff}} = \mathbb{E}_{t, x_0, \epsilon} \left[ \| x_t - x_0 \|^2 \right],
\end{equation}
where $\epsilon$ denotes noise drawn from a standard Gaussian distribution. Minimizing this loss allows the model to iteratively refine the generated image, progressively reducing noise and bringing it closer to the angiography.
%-------------------------------------------------------------------------
\subsection{Vascular Geometric State Space Model}\label{sec:ssm}
The geometric structure of vessels is crucial for clinical diagnosis. To emphasize these features in angiography synthesis, we map the volumetric properties of vascular geometry into a three-dimensional state space during reverse diffusion. Additionally, we introduce embedding representations via vision embedder and cross-slice attention, enabling effective geometric synthesis of angiography.

The vascular tree-state space model takes two inputs. Firstly, Sec.~\ref{sec:embedder} generates the embedding representations of the non-angiography mask $M$ and tokens $T$, which guide the synthesis process. Secondly, the angiographic latent representation $z^{Noise,t}_{\text{angio.}}$ from forward diffusion is converted into multiple tokens as the other input, which are enhanced with positional embeddings to maintain spatial context.

The inputs to each vascular tree-state space model include the latent representation $z^{Noise,t}_{\text{angio.}}$ of angiography and the embedding representations of the non-angiographic mask $M$ and tokens $T$. A linear mapping layer converts each local feature into fixed-dimensional embeddings, followed by deep convolution for alignment. The mamba block adapts to conditional inputs, and long skip connections provide feedback to avoid numerical issues in multi-layer processing. Features after layer normalization are connected residually to form the final output. 

\begin{figure}
  \centering
  \includegraphics[width=\linewidth]{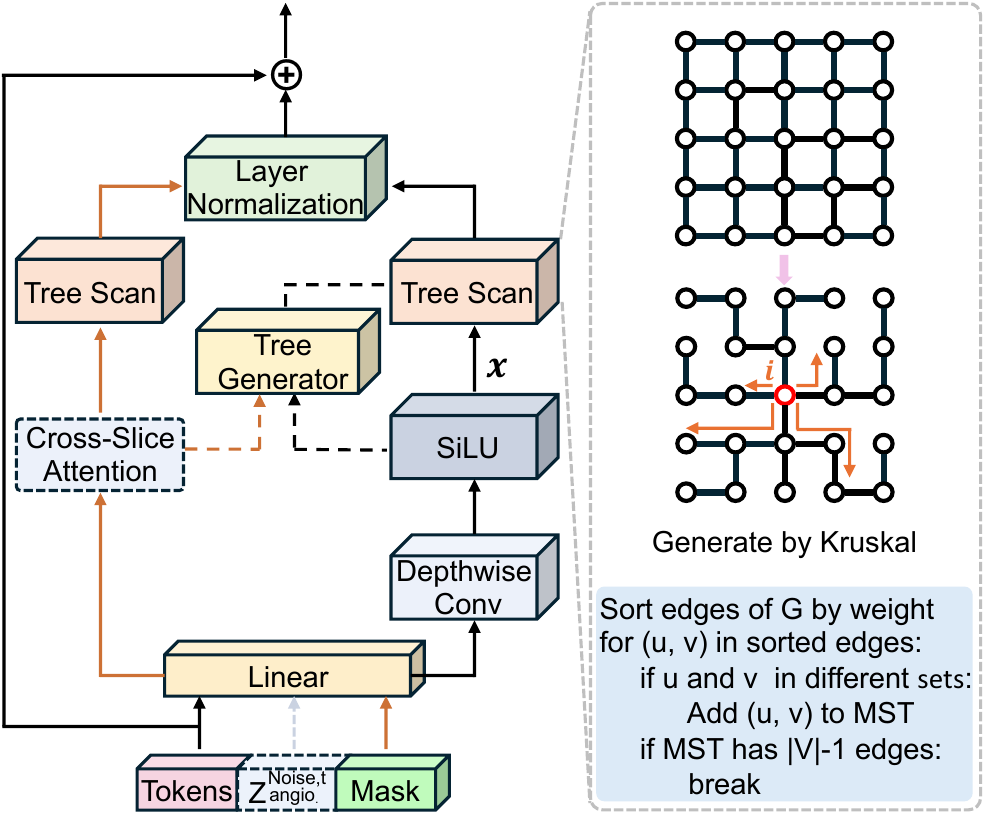} 
  \caption{\textbf{Vascular Tree-state space architecture.} A dynamic tree is constructed from vessel features and spatial relationships, with embeddings guided by prior slice information.}
  \label{fig:mamba}
  \vspace{-0.6cm}
\end{figure}

The vascular tree-state space model, illustrated in Fig.~\ref{fig:mamba}, combines the vascular tree scanning algorithm presented in Sec.~\ref{sec:Vascular Tree Scanning} with the cross-slice attention mechanism discussed in Sec.~\ref{sec:cross-slice}, highlighting the significance of different regions within the slice.

% \textbf{Vascular Tree Scanning Algorithm.}\label{Vascular Tree Scanning}
\subsubsection{Vascular Tree Scanning Algorithm}\label{sec:Vascular Tree Scanning}
The tree scanning algorithm constructs an input-adaptive tree topology to transmit and aggregate features efficiently. Input a feature map $X = \{x_i\}_{i=1}^L$, where $L$ is the sequence length, an undirected, $m$-connected graph $G = (V, E)$ is created, with $m = 4$ ensuring each node is connected to its four neighbors. The edges of this graph are weighted using feature dissimilarity, typically measured by cosine distance. Kruskal’s algorithm is then applied to prune the graph into a Minimum Spanning Tree $G_T$, which preserves spatial relationships while minimizing edge weights. The specific algorithm flow is shown in Algorithm~\ref{alg:tree_scanning}.

State propagation is performed by treating each node as a root and aggregating features from neighboring nodes to capture long-range dependencies. This process is formalized by a state transition matrix and dynamic programming, ensuring information flows both top-down and bottom-up across the tree. At each node $i$, feature aggregation is given by $h[i] = \sum_{j \in \text{Neigh}(i)}S(E[i,j]) \cdot B[j] \cdot X[j]$, where $\text{Neigh}(i)$ is the set of neighboring nodes, $S(E[i,j])$ represents the edge weight, $B[j]$ is the input transformation matrix, and $X[j]$ is the input feature. This method preserves spatial structure and enhances feature representation, enabling efficient aggregation of long-range dependencies.
\begin{algorithm}[htbp]
\caption{Vascular Tree Scanning Algorithm}
\label{alg:tree_scanning}
\SetKwInOut{Require}{Require}
\SetKwInOut{Ensure}{Ensure}
\Require{$X = \{x_i\}_{i=1}^L$, $Y$}
\Ensure{$output$, gradients}

\BlankLine
Initialize: 
$G \gets \text{initializeGraph}(V, E)$,
$G_T \gets \text{kruskalMST}(G)$,
$h[i] = 0, \forall i \in G_T$\;

\For{$i = 1$ \textbf{to} $L$}{
    \ForEach{$j \in \text{getNeighbors}(i, 4)$}{
        $\text{addEdge}(G, i, j, \text{cosineDist}(X[i], X[j]))$\;
    }
}
Update: 
$h[i] = \sum_{j \in \text{Neigh}(i)} S(E[i,j]) \cdot B[j] \cdot X[j], \forall i \in G_T$\;

Compute: 
$output = C \cdot (A \cdot h + D \cdot X)$,
$\eta_i \gets \frac{\partial \mathcal{L}}{\partial h_i}$\;

\For{$i \in G_T$ (Bottom-up \& Top-down)}{
    Update: $\eta_i \gets \bar{B}_i \eta_i + \sum_{j \in \text{Children}(i)} \eta_j \bar{A}_j$\;
    Propagate: $\eta_j \gets \bar{B}_j \eta_i, \quad \forall j \in \text{Children}(i)$\;
}
Calculate: $\mathcal{L}_{\text{scan}} = \sum_{i} \frac{1}{2} \| \eta_i \|^2$\;

\Return{$output$, gradients}\;
\end{algorithm}

Tree scan loss is computed via backpropagation over a Minimum Spanning Tree (MST), consisting of two stages: aggregation and state propagation. In the aggregation stage, the gradient for each node $i$ is computed as the partial derivative of the loss with respect to its hidden state $h_i$.
\begin{equation}
\eta_i = \frac{\partial \text{Loss}}{\partial h_i},
\end{equation}
\vspace{-0.3cm}
\begin{equation}
\eta_i = \bar{B}_i \frac{\partial \text{Loss}}{\partial h_i} + \sum_{\forall j \in \{t \mid \text{Par}(t) = i\}} \eta_j \bar{A}_j,
\end{equation}
where $\eta_i$ represents the gradient of the hidden state $h_i$, $\bar{B}_i$ is the input matrix for node $i$, $\bar{A}_j$ is the state matrix for the child node $j$, $\text{Par}(t)$ denotes the parent of node $t$.

In the state propagation stage, the gradients propagate from the root node to the leaf nodes. The gradients for each node $i$ and its child node $j$ are updated as follows:
\begin{equation}
\eta_i = \bar{A}_i \eta_i, \quad \eta_j = \bar{B}_j \eta_i.
\end{equation}
The tree scan loss is then defined as the sum of the squared magnitudes of the gradients across all nodes:
\begin{equation}
\mathcal{L}_{\text{scan}} = \sum_{i} \frac{1}{2} \| \eta_i \|^2.
\end{equation} 

The overall loss function is composed of the InfoNCE loss from the vision embedder, the denoising loss from the denoising module, and the tree scan loss from the vascular scanning in the tree-state space. It can be expressed as:
\begin{equation}
\mathcal{L}_{\text{total}} = \alpha \cdot \mathcal{L}_{\text{InfoNCE}} + \beta \cdot \mathcal{L}_{\text{diff}} + \gamma \cdot \mathcal{L}_{\text{scan}},
\end{equation}

\noindent where $\alpha$, $\beta$, and $\gamma$ are the respective weights assigned to each component of the loss function.

\begin{table*}[ht]
\centering
\begin{tabular}{c|cc|cc|cc|cc|cc}
\hline
\multirow{3}{*}{Methods} & \multicolumn{6}{c|}{ITKTubeTK}                                                                                                                           & \multicolumn{2}{c|}{ISICDM 2020}     & \multicolumn{2}{c}{Topcow 2024}    \\ \cline{2-11} 
                         & \multicolumn{2}{c|}{T1-Flash}                            & \multicolumn{2}{c|}{T2}                                  & \multicolumn{2}{c|}{T1-MPRAGE}      & \multicolumn{2}{c|}{CT}             & \multicolumn{2}{c}{CTA}             \\ \cline{2-11} 
                         & \centering PSNR         & SSIM                              & PSNR      & SSIM                              & PSNR                    & SSIM                     & PSNR                 & SSIM                 & PSNR                 & SSIM                 \\ \hline
SAGAN~\cite{zhang2019self}                   & 27.85        & 0.9184                            & 28.16     & 0.9254                            & 27.96                   & 0.9234                   & 26.42                & 0.8921               & 25.32                & 0.9018               \\
cGAN~\cite{dar2019image}
& 28.45        & 0.9248                            & 28.73     & 0.9289                            & 28.68                   & 0.9264                   & 27.87                & 0.9163               & 25.58                & 0.9142               \\ \hline
SynDiff~\cite{ozbey2023unsupervised}
& 28.98        & \underline{0.9446}                            & 29.06     & \textbf{0.9482}                            & 29.12                   & \textbf{0.9516}                   & 28.56                & 0.9321               & 26.13                & \underline{0.9317}               \\
Fast-DDPM~\cite{jiang2024fast}               & 26.92        & 0.8958                            & 27.51     & 0.9024                            & 27.37                   & 0.8984                   & 27.38                & 0.9128               & 24.15                & 0.8924               \\ 
DiffMa~\cite{wang2024soft}                  & \underline{29.12}        & 0.6936                            & \underline{29.36}     & 0.7174                            & \underline{29.28}                   & 0.7058                   & \underline{28.94}                & \underline{0.9428}               & \underline{26.76}                & 0.7286               \\ \hline
VasTSD                 & \textbf{29.86}        & \textbf{0.9478}                            & \textbf{29.72}     & \underline{0.9384}                            & \textbf{29.92}                   & \underline{0.9454}                   & \textbf{29.48}                & \textbf{0.9562}               & \textbf{27.36}                & \textbf{0.9372}               \\ \hline
\end{tabular}
\caption{\textbf{Comparison of modal conversion methods for angiography synthesis.} PSNR $\uparrow$ and SSIM $\uparrow$ results across different datasets. The best and second-best results are highlighted in \textbf{bold} and \underline{underlined}.}
\label{tab:result_comparison}
\vspace{-0.6cm}
\end{table*}

\subsubsection{Cross-Slice Attention}\label{sec:cross-slice}
Traditional State Space Models (SSMs) map input sequences $\mathbf{u}(t)$ to output sequences $\mathbf{y}(t)$. Recent advancements, such as S4~\cite{gu2021efficiently}, S5~\cite{smith2022simplified}, and S6~\cite{gu2023mamba}, have improved the efficiency of SSMs. For instance, S6 introduces a dynamic selection mechanism in the $A$-matrix that adapts based on the input content. However, the serialization method in S6 processes each slice independently, making it difficult to capture the 3D vascular structure.

To overcome this limitation, this paper proposes a cross-slice attention mechanism that leverages pre-trained non-angiography embeddings and spatial constraints from slice masks. Within each slice mask, weights are high in vascular regions and low in non-vascular areas. Each vascular region is represented as a feature vector, and the similarity between across slices is computed using cosine similarity:
\begin{equation}
S_{ij} = \frac{(h_i \cdot h_j)}{\|h_i\| \|h_j\|},
\end{equation}
where $h_i$ and $h_j$ are the feature vectors for slices $i$ and $j$. To this end, the cross-slice attention coefficient $\alpha_{ij}$ represents the attention from node $i$ to node $j$ across slices:
\begin{equation}
\alpha_{ij} = \frac{\exp(\text{LReLU}(S_{ij} + w \cdot \text{mask}_i \cdot \text{mask}_j))}{\sum_k \exp(\text{LReLU}(S_{ik} + w \cdot \text{mask}_i \cdot \text{mask}_k))},
\end{equation}
where $w$ is a hyperparameter controlling mask influence. LReLU represents the LeakyReLU activation function. This cross-slice attention captures the spatial continuity of vascular structures, enabling dynamic state transition matrix construction for 3D vascular synthesis. 
\section{Experiments}
%-------------------------------------------------------------------------
\subsection{Experimental Details}
\textbf{Dataset.} This paper conducts several experiments on brain and lung vessels. The brain vascular data include the ITKTubeTK dataset~\cite{gungor2023deq} and the Topcow2024 dataset~\cite{yang2023benchmarking}. The pulmonary vessels from ISICDM 2020 dataset~\cite{tan2021automated, han20233d}. All datasets undergo specific preprocessing steps, as detailed in the supplementary material.

\noindent\textbf{Implementation Details.} The vision embedder is pre-trained on an NVIDIA A100 GPU using non-angiography, with the Variational Auto-Encoder (VAE) employing an exponential moving average (EMA) mode. During the training of the diffusion model, two A100 GPUs are used with the learning rate of $1 \times 10^{-4}$ and batch size of 256.

%-------------------------------------------------------------------------
\noindent\textbf{Metrics.} 
In this paper, we evaluate the quality of synthetic vascular 2D slices generated by different methods using two quantitative metrics: Peak Signal-to-Noise Ratio (PSNR) and Structural Similarity Index (SSIM). The Dice and Jaccard scores evaluate the 3D vascular morphological similarity, while the Connectivity score measures spatial connectivity. Details are provided in the supplementary.
\begin{figure*}
  \centering
  \includegraphics[width=\textwidth]{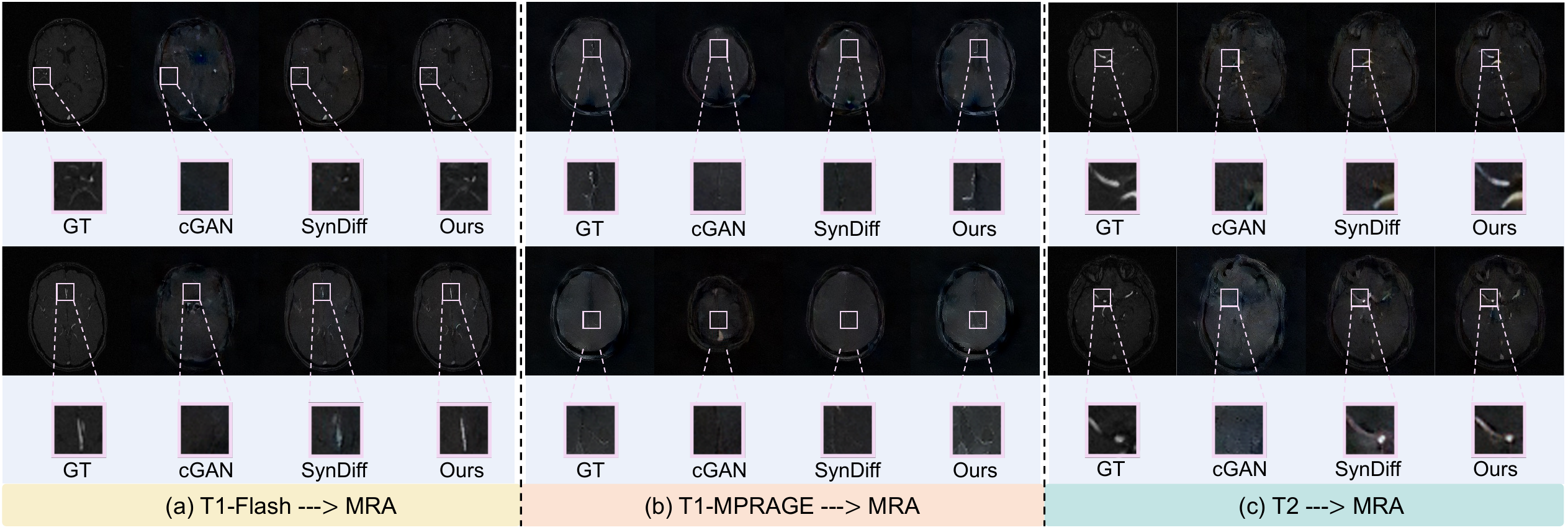} 
  \caption{\textbf{Comparison of 2D angiographic slice modality transformation and ours on the ITKTubeTK dataset.} From left to right: MRA results corresponding to T1-Flash, T1-MPRAG, and T2, with comparisons between the cGAN~\cite{dar2019image}, SynDiff~\cite{ozbey2023unsupervised}, and our method.}
  \label{fig:result1}
  \vspace{-0.3cm}
\end{figure*}
\begin{figure}
  \centering
  \includegraphics[width=0.46\textwidth]{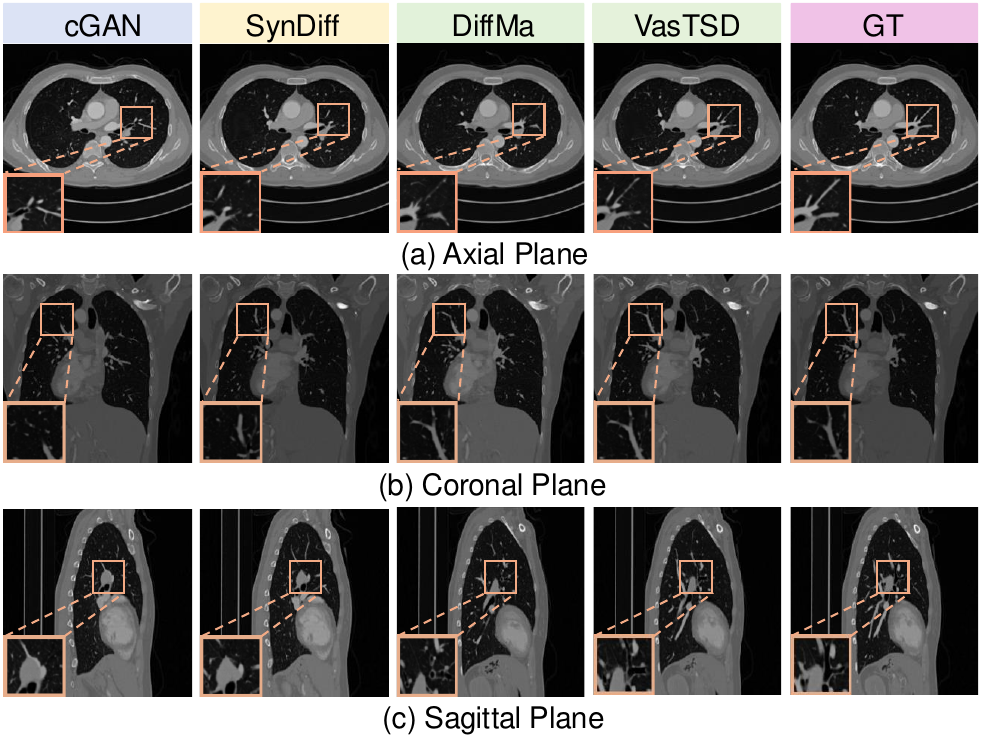} 
  \caption{\textbf{Synthesized CTA from the pulmonary artery (non-angiographic) CTs, by cGAN~\cite{dar2019image}, SynDiff~\cite{ozbey2023unsupervised}, DiffMa~\cite{wang2024soft}, and our method.} From top to bottom, visualized from three different anatomical coordinate systems of the body.}
  \label{fig:result_cta}
  \vspace{-0.6cm}
\end{figure}
\begin{figure*}
  \centering
  \includegraphics[width=\textwidth]{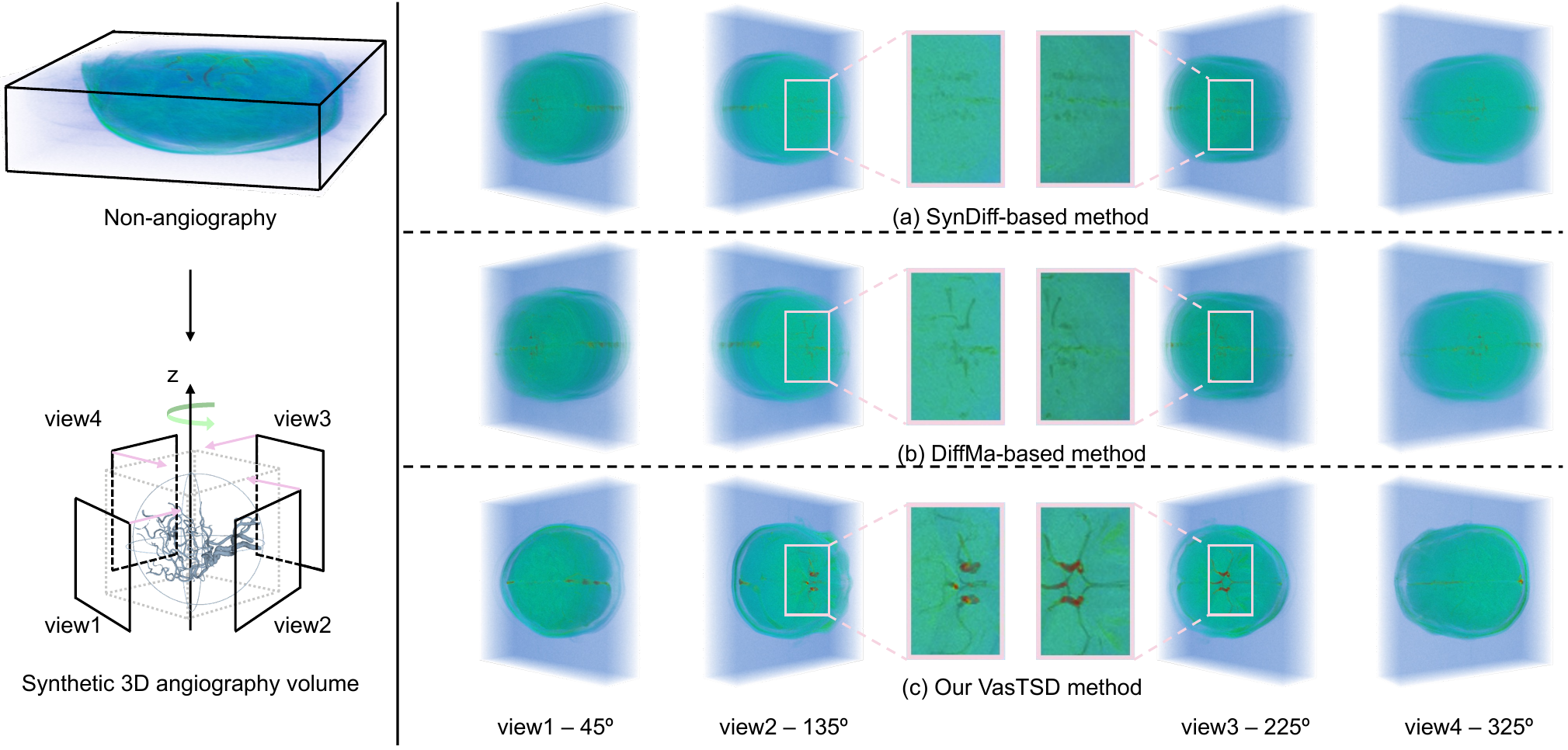} 
  \caption{\textbf{Comparison of synthetic angiographic 3D volumes from four perspectives.} \textit{Left:} Schematic of the 3D angiography synthesis process. \textit{Right:} To enhance the visualization of vascular regions, the original data is processed. The first and second rows are based on SynDiff~\cite{ozbey2023unsupervised} and DiffMa~\cite{wang2024soft} cross-modal transformation methods, while the third row presents the synthetic outcomes of our approach.}
  \label{fig:result3d}
  \vspace{-0.5cm}
\end{figure*}
\begin{figure}
  \centering
  \includegraphics[width=0.47\textwidth]{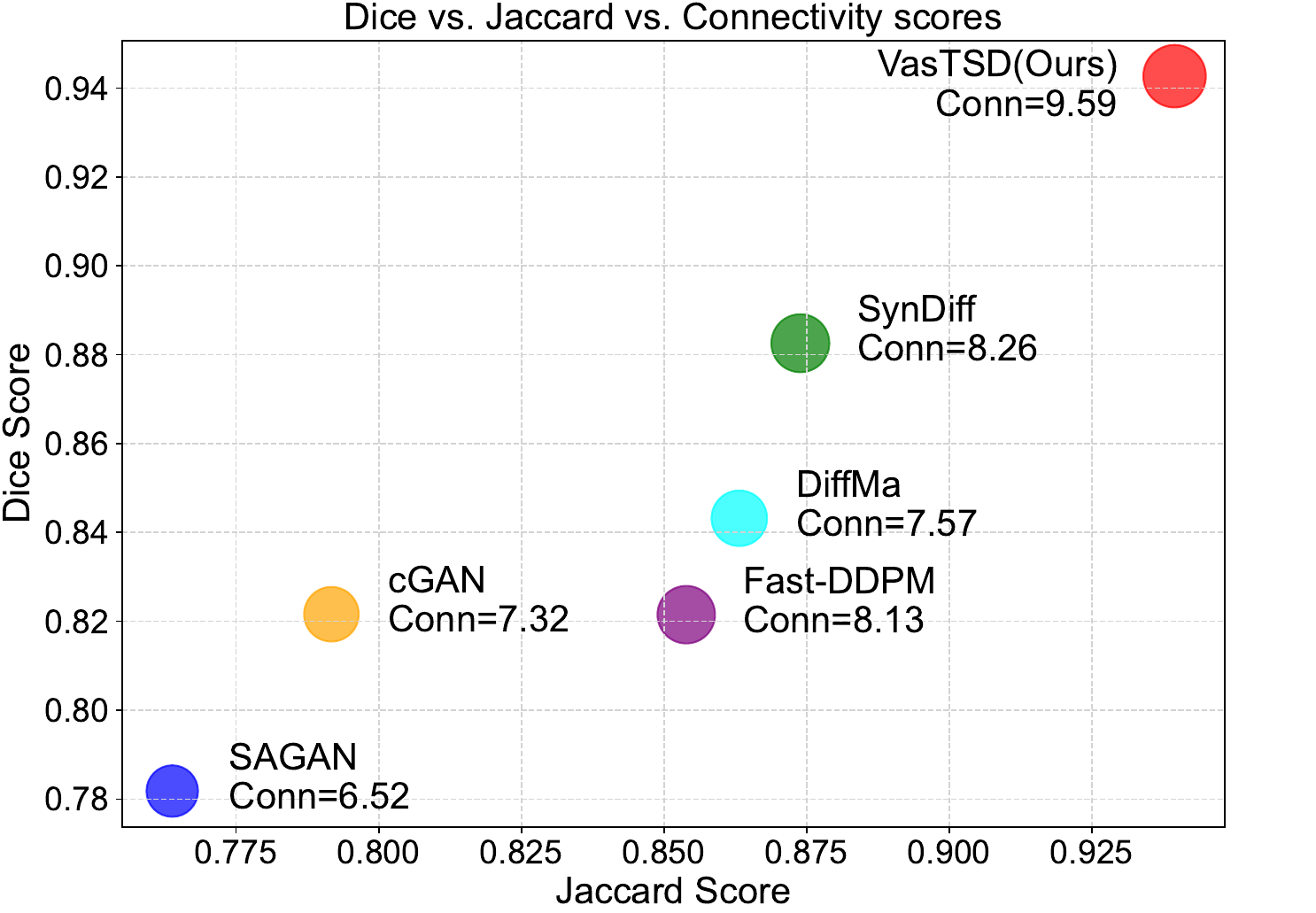} 
  \caption{\textbf{3D vascular morphology and connectivity evaluation.} Higher Dice/Jaccard scores (0–1) indicate greater similarity to real blood vessels, while a higher connectivity score (0–10) reflects better spatial connectivity, smoothness, and structural quality.}
  \label{fig:result_3d2}
  \vspace{-0.3cm}
\end{figure}
%-------------------------------------------------------------------------
\subsection{Comparisons}
This work evaluates the effectiveness of the proposed 3D Tree-state Space Diffusion (VasTSD) model through experiments on both 2D vascular slices and 3D synthetic angiography volumes applied to brain and lung vasculature.

\noindent\textbf{2D Vascular Slice Evaluation.}
To assess the microstructure of synthetic angiography, we compare VasTSD with classic modality conversion methods, including GAN-based SAGAN~\cite{zhang2019self}, cGAN~\cite{dar2019image}, and several diffusion-based methods, such as SynDiff~\cite{ozbey2023unsupervised}, Fast-DDPM~\cite{jiang2024fast}, and DiffMa~\cite{wang2024soft}. All input data are aligned to ensure fairness in comparison. We visualize synthetic angiography results generated from non-angiographic MRI sequences in the ITKTubeTK dataset in Fig.~\ref{fig:result1}. As shown in the red-marked part, the VasTSD obviously outperforms these modality conversion approaches, particularly in synthesizing small vessels and capturing vascular curvature. Based on the ISICDM 2020 pulmonary artery dataset, as shown in Fig.~\ref{fig:result_cta}, we visualize the 2D slice performance of synthesizing CTA from CT in the body coordinate system. In terms of the branching structure of the vessels, we observe that VasTSD can achieve more accurate and continuous blood branches. The synthesis performance of 2D vascular slices is summarized in Tab.~\ref{tab:result_comparison}. It can be observed that our method achieves state-of-the-art (SOTA) performance across most experiments. Compared to other methods, our approach achieves the highest PSNR across all datasets, with the maximum value reaching 29.92. Although the SSIM for T2 and T1-MPRAGE are slightly lower than those achieved by SynDiff, it is important to note that SynDiff is a 2D approach. In contrast, our method focuses on the connectivity of vascular structures, which is beneficial for disease diagnosis and holds the potential for improved clinical utility.

\noindent\textbf{3D Vascular Volume Evaluation.}
The spatial topology of vessels is crucial for diagnosing vascular diseases and surgical planning. This work evaluates VasTSD from the perspective of 3D vascular volume. Due to the lack of previous research on synthetic 3D angiography, the comparative method is based on existing 2D vascular synthesis methods by stacking 2d slices according to their spatial positions. Inspired by multi-view NeRF~\cite{wen2024cad}, we visualize the 3D vascular volume from four perspectives, as shown in Fig.~\ref{fig:result3d}. Compared to SynDiff and DiffMa, VasTSD generates a more integrated 3D vasculature, demonstrating superior integrity. To quantify the connectivity of the synthetic 3D vessels and their morphological similarity to real blood vessels, the evaluation results for Dice, Jaccard, and Connectivity Scores are presented in Fig.~\ref{fig:result_3d2}. The results indicate that VasTSD excels in these metrics, demonstrating its superior geometric continuity across vascular slices. Statistical analyses of 3D volumes are provided in the supplementary.

\subsection{Ablation Studies}
We conduct ablation studies to analyze the contribution of components in the VasTSD to its performance. Specifically, we investigate the impact of the vision embedder, the scanning of the state space, and the loss function. The ablation study on the loss is provided in the supplementary materials.

\noindent\textbf{Ablation Comparison of Vision Embedder.} In this work, to achieve consistent vascular geometry across different slices, we design a pre-trained vision embedder to capture conditional information $m, t$, which enhances the reverse diffusion process of the model. Here we test two other configurations for comparison: one without the pre-trained mechanism and another where BioMedCLIP~\cite{zhang2023biomedclip} is used to generate the conditional information directly. As shown in Tab.~\ref{tab:abla_embedder}, the pre-trained vision embedder representation significantly improves the accuracy of the generated angiography. Compared to the BioMedCLIP vision embedder, which is pre-trained on large-scale medical datasets. Our method demonstrates superior performance in PSNR and SSIM on 2D slices synthesized for multi-modal and anatomical blood vessels. This further underscores the effectiveness of our approach for various types of vessels.
\begin{table}[ht]
\centering
\resizebox{\linewidth}{!}{
\begin{tabular}{c|c|c|c|c}
\hline
\multirow{2}{*}{Methods} & \multicolumn{2}{c|}{ITKTubeTK (T1-Flash)} & \multicolumn{2}{c}{ISICDM 2020} \\ \cline{2-5} 
                         & \makebox[1.5cm][c]{PSNR} & \makebox[1.5cm][c]{SSIM} & \makebox[1.5cm][c]{PSNR} & \makebox[1.5cm][c]{SSIM} \\ \hline
w/o pre-training         & 26.36          & 0.9172         & 27.57          & 0.8993         \\ \hline
BioMedCLIP               & 28.68          & 0.9286         & 28.53          & 0.9296         \\ \hline
Ours                     & \textbf{29.86}          & \textbf{0.9478}         & \textbf{29.48}          & \textbf{0.9562}         \\ \hline
\end{tabular}}
\caption{\textbf{Compare the effects of different vision embedders.} PSNR $\uparrow$ and SSIM $\uparrow$ results across the dataset.}
\label{tab:abla_embedder}
\vspace{-0.6cm}
\end{table}

\noindent\textbf{Ablation Comparison of Vascular State Space Scans.} In this work, we design a vascular state space scanning method based on the physical principles of vascular geometry, so we investigate the impact of different vascular serialization methods on vascular continuity synthesis. 
We evaluate the synthesized vascular angiograms and corresponding model parameter counts on the ITKTubeTK dataset, with the results shown in Tab.~\ref{tab:scan_result}. Compared with two classic state-space serialization methods, our approach achieves higher angiogram synthesis accuracy, especially with greater improvement in structural similarity. At the same time, our model employs the fewest number of parameters. 
\begin{table}[ht]
\centering
\resizebox{\linewidth}{!}{
\begin{tabular}{c|cc|cc|c}
\hline
\multirow{2}{*}{Backbone} & \multicolumn{2}{c|}{T1-Flash} & \multicolumn{2}{c|}{T2} & \multirow{2}{*}{Params(M)} \\ \cline{2-5}
                          & \multicolumn{1}{c|}{PSNR}         & SSIM          & \multicolumn{1}{c|}{PSNR}      & SSIM       &                             \\ \hline
Tree-scan                 & \multicolumn{1}{c|}{\textbf{29.86}}        & \textbf{0.9478}        & \multicolumn{1}{c|}{\textbf{29.72}}     & \textbf{0.9384}     & \textbf{79.24}                       \\ \hline
Zigzag-scan-B             & \multicolumn{1}{c|}{26.53}        & 0.8216        & \multicolumn{1}{c|}{26.25}     & 0.8138     & 133.80                       \\ \hline
Local-scan-S              & \multicolumn{1}{c|}{28.96}        & 0.8729        & \multicolumn{1}{c|}{28.54}     & 0.8726     & 81.83  \\ \hline
\end{tabular}}
\caption{\textbf{Ablation effects of different state space serialization methods.} The best results are highlighted in \textbf{bold}. PSNR $\uparrow$ and SSIM $\uparrow$ results across the ITKTubeTK (T1-Flash).}
\label{tab:scan_result}
\vspace{-0.6cm}
\end{table}

% \input{sec/5_discussion}
%-------------------------------------------------------------------------
\section{Conclusions}
This work introduces a method VasTSD %, a 3D vascular tree-state space diffusion model 
for angiography synthesis, which can be used to reduce patient exposure to contrast agents in clinical settings. VasTSD dynamically constructs vascular tree topology and integrates it with a diffusion-based generative model, ensuring continuity in synthesized 3D vessels, working for different modalities. 

%-------------------------------------------------------------------------
\noindent\textbf{Limitations.}
The anisotropy inherent in medical data acquisition may lead to varying image resolution, texture, and tissue response properties in different directions due to physical characteristics and differences in tissue structures across various orientations. Our experiments based on different slice orientations also exhibit some degree of independence. 
Further explorations of this problem are necessary and meaningful. 
\paragraph{Acknowledgements. } 
This work is supported in part by the NSFC (62325211, 62132021, 62372457), the Major Program of Xiangjiang Laboratory (23XJ01009), Young Elite Scientists Sponsorship Program by CAST (2023QNRC001), the Natural Science Foundation of Hunan Province of China (2022RC1104) and the NUDT Research Grants (ZK22-52).

{
    \small
    \bibliographystyle{ieeenat_fullname}
    \bibliography{main}

\begin{thebibliography}{51}
\providecommand{\natexlab}[1]{#1}
\providecommand{\url}[1]{\texttt{#1}}
\expandafter\ifx\csname urlstyle\endcsname\relax
  \providecommand{\doi}[1]{doi: #1}\else
  \providecommand{\doi}{doi: \begingroup \urlstyle{rm}\Url}\fi

\bibitem[Armanious et~al.(2020)Armanious, Jiang, Fischer, K{\"u}stner, Hepp, Nikolaou, Gatidis, and Yang]{armanious2020medgan}
Karim Armanious, Chenming Jiang, Marc Fischer, Thomas K{\"u}stner, Tobias Hepp, Konstantin Nikolaou, Sergios Gatidis, and Bin Yang.
\newblock Medgan: Medical image translation using gans.
\newblock \emph{Computerized medical imaging and graphics}, 79:\penalty0 101684, 2020.

\bibitem[Chung and Ye(2022)]{chung2022score}
Hyungjin Chung and Jong~Chul Ye.
\newblock Score-based diffusion models for accelerated mri.
\newblock \emph{Medical image analysis}, 80:\penalty0 102479, 2022.

\bibitem[Dar et~al.(2019)Dar, Yurt, Karacan, Erdem, Erdem, and Cukur]{dar2019image}
Salman~UH Dar, Mahmut Yurt, Levent Karacan, Aykut Erdem, Erkut Erdem, and Tolga Cukur.
\newblock Image synthesis in multi-contrast mri with conditional generative adversarial networks.
\newblock \emph{IEEE transactions on medical imaging}, 38\penalty0 (10):\penalty0 2375--2388, 2019.

\bibitem[Deshpande et~al.(2024)Deshpande, {\"O}zbey, Li, Anastasio, and Brooks]{deshpande2024assessing}
Rucha Deshpande, Muzaffer {\"O}zbey, Hua Li, Mark~A Anastasio, and Frank~J Brooks.
\newblock Assessing the capacity of a denoising diffusion probabilistic model to reproduce spatial context.
\newblock \emph{IEEE Transactions on Medical Imaging}, 2024.

\bibitem[Gao et~al.(2023)Gao, Li, Zhang, Zhang, and Shan]{gao2023corediff}
Qi Gao, Zilong Li, Junping Zhang, Yi Zhang, and Hongming Shan.
\newblock Corediff: Contextual error-modulated generalized diffusion model for low-dose ct denoising and generalization.
\newblock \emph{IEEE Transactions on Medical Imaging}, 2023.

\bibitem[Gong et~al.(2024)Gong, Kang, Wang, Wan, and Li]{gong2024nnmamba}
Haifan Gong, Luoyao Kang, Yitao Wang, Xiang Wan, and Haofeng Li.
\newblock nnmamba: 3d biomedical image segmentation, classification and landmark detection with state space model.
\newblock \emph{arXiv preprint arXiv:2402.03526}, 2024.

\bibitem[Goodfellow et~al.(2020)Goodfellow, Pouget-Abadie, Mirza, Xu, Warde-Farley, Ozair, Courville, and Bengio]{goodfellow2020generative}
Ian Goodfellow, Jean Pouget-Abadie, Mehdi Mirza, Bing Xu, David Warde-Farley, Sherjil Ozair, Aaron Courville, and Yoshua Bengio.
\newblock Generative adversarial networks.
\newblock \emph{Communications of the ACM}, 63\penalty0 (11):\penalty0 139--144, 2020.

\bibitem[Gu and Dao(2023)]{gu2023mamba}
Albert Gu and Tri Dao.
\newblock Mamba: Linear-time sequence modeling with selective state spaces.
\newblock \emph{arXiv preprint arXiv:2312.00752}, 2023.

\bibitem[Gu et~al.(2021)Gu, Goel, and R{\'e}]{gu2021efficiently}
Albert Gu, Karan Goel, and Christopher R{\'e}.
\newblock Efficiently modeling long sequences with structured state spaces.
\newblock \emph{arXiv preprint arXiv:2111.00396}, 2021.

\bibitem[G{\"u}ng{\"o}r et~al.(2023)G{\"u}ng{\"o}r, Askin, Soydan, Top, Saritas, and {\c{C}}ukur]{gungor2023deq}
Alper G{\"u}ng{\"o}r, Baris Askin, Damla~Alptekin Soydan, Can~Bari{\c{s}} Top, Emine~Ulku Saritas, and Tolga {\c{C}}ukur.
\newblock Deq-mpi: A deep equilibrium reconstruction with learned consistency for magnetic particle imaging.
\newblock \emph{IEEE Transactions on Medical Imaging}, 43\penalty0 (1):\penalty0 321--334, 2023.

\bibitem[Han et~al.(2023)Han, He, Zheng, Li, and Ma]{han20233d}
Jiachen Han, Naixin He, Qiang Zheng, Lin Li, and Chaoqing Ma.
\newblock 3d pulmonary vessel segmentation based on improved residual attention u-net.
\newblock \emph{Medicine in Novel Technology and Devices}, 20:\penalty0 100268, 2023.

\bibitem[He et~al.(2024)He, Peng, Yi, Wu, and Wang]{he2024multi}
Yuhong He, Long Peng, Qiaosi Yi, Chen Wu, and Lu Wang.
\newblock Multi-scale representation learning for image restoration with state-space model.
\newblock \emph{arXiv preprint arXiv:2408.10145}, 2024.

\bibitem[Ho et~al.(2020)Ho, Jain, and Abbeel]{ho2020denoising}
Jonathan Ho, Ajay Jain, and Pieter Abbeel.
\newblock Denoising diffusion probabilistic models.
\newblock \emph{Advances in neural information processing systems}, 33:\penalty0 6840--6851, 2020.

\bibitem[Hu et~al.(2024)Hu, Baumann, Gui, Grebenkova, Ma, Fischer, and Ommer]{hu2024zigma}
Vincent~Tao Hu, Stefan~Andreas Baumann, Ming Gui, Olga Grebenkova, Pingchuan Ma, Johannes~S Fischer, and Bj{\"o}rn Ommer.
\newblock Zigma: A dit-style zigzag mamba diffusion model.
\newblock \emph{arXiv preprint arXiv:2403.13802}, 2024.

\bibitem[Huang et~al.(2024)Huang, Pei, You, Wang, Qian, and Xu]{huang2024localmamba}
Tao Huang, Xiaohuan Pei, Shan You, Fei Wang, Chen Qian, and Chang Xu.
\newblock Localmamba: Visual state space model with windowed selective scan.
\newblock \emph{arXiv preprint arXiv:2403.09338}, 2024.

\bibitem[Jiang et~al.(2024)Jiang, Imran, Ma, Zhang, Zhou, Liang, Gong, and Shao]{jiang2024fast}
Hongxu Jiang, Muhammad Imran, Linhai Ma, Teng Zhang, Yuyin Zhou, Muxuan Liang, Kuang Gong, and Wei Shao.
\newblock Fast denoising diffusion probabilistic models for medical image-to-image generation.
\newblock \emph{arXiv preprint arXiv:2405.14802}, 2024.

\bibitem[Kamran et~al.(2021)Kamran, Hossain, Tavakkoli, and Zuckerbrod]{kamran2021attention2angiogan}
Sharif~Amit Kamran, Khondker~Fariha Hossain, Alireza Tavakkoli, and Stewart~Lee Zuckerbrod.
\newblock Attention2angiogan: Synthesizing fluorescein angiography from retinal fundus images using generative adversarial networks.
\newblock In \emph{2020 25th International Conference on Pattern Recognition (ICPR)}, pages 9122--9129. IEEE, 2021.

\bibitem[Kazerouni et~al.(2023)Kazerouni, Aghdam, Heidari, Azad, Fayyaz, Hacihaliloglu, and Merhof]{kazerouni2023diffusion}
Amirhossein Kazerouni, Ehsan~Khodapanah Aghdam, Moein Heidari, Reza Azad, Mohsen Fayyaz, Ilker Hacihaliloglu, and Dorit Merhof.
\newblock Diffusion models in medical imaging: A comprehensive survey.
\newblock \emph{Medical Image Analysis}, 88:\penalty0 102846, 2023.

\bibitem[Kingma(2013)]{kingma2013auto}
Diederik~P Kingma.
\newblock Auto-encoding variational bayes.
\newblock \emph{arXiv preprint arXiv:1312.6114}, 2013.

\bibitem[Koch et~al.(2024)Koch, Aydin, Hilbert, Rieger, Tanioka, Ishida, and Frey]{koch2024cross}
Alexander Koch, Orhun~Utku Aydin, Adam Hilbert, Jana Rieger, Satoru Tanioka, Fujimaro Ishida, and Dietmar Frey.
\newblock Cross-modality image synthesis from tof-mra to cta using diffusion-based models.
\newblock \emph{arXiv preprint arXiv:2409.10089}, 2024.

\bibitem[Kshirsagar et~al.(2024)Kshirsagar, McNulty, Taji, So, Chong, Theriault-Lauzier, Wisniewski, and Shrimohammadi]{kshirsagar2024generative}
Jay Kshirsagar, John McNulty, Bahareh Taji, Derek So, Aun-Yeong Chong, Pascal Theriault-Lauzier, AJ Wisniewski, and Shervin Shrimohammadi.
\newblock Generative ai-assisted novel view synthesis of coronary arteries for angiography.
\newblock In \emph{2024 IEEE International Symposium on Medical Measurements and Applications (MeMeA)}, pages 1--6. IEEE, 2024.

\bibitem[Liu et~al.(2024{\natexlab{a}})Liu, Li, and Yuan]{liu2024diffrect}
Xinyu Liu, Wuyang Li, and Yixuan Yuan.
\newblock Diffrect: Latent diffusion label rectification for semi-supervised medical image segmentation.
\newblock In \emph{International Conference on Medical Image Computing and Computer-Assisted Intervention}, 2024{\natexlab{a}}.

\bibitem[Liu et~al.(2024{\natexlab{b}})Liu, Tian, Zhao, Yu, Xie, Wang, Ye, and Liu]{liu2024vmamba}
Yue Liu, Yunjie Tian, Yuzhong Zhao, Hongtian Yu, Lingxi Xie, Yaowei Wang, Qixiang Ye, and Yunfan Liu.
\newblock Vmamba: Visual state space model.
\newblock \emph{arXiv preprint arXiv:2401.10166}, 2024{\natexlab{b}}.

\bibitem[Lyu et~al.(2023{\natexlab{a}})Lyu, Fu, Yang, Xiong, Duan, Duan, Wang, Xing, Zhang, Lin, et~al.]{lyu2023generative}
Jinhao Lyu, Ying Fu, Mingliang Yang, Yongqin Xiong, Qi Duan, Caohui Duan, Xueyang Wang, Xinbo Xing, Dong Zhang, Jiaji Lin, et~al.
\newblock Generative adversarial network--based noncontrast ct angiography for aorta and carotid arteries.
\newblock \emph{Radiology}, 309\penalty0 (2):\penalty0 e230681, 2023{\natexlab{a}}.

\bibitem[Lyu et~al.(2023{\natexlab{b}})Lyu, Tan, Lipford, Niu, Zapadka, Lack, Clemente, Whitlow, and Wang]{lyu2023head}
Qing Lyu, Josh Tan, Megan~E Lipford, Chuang Niu, Micheal~E Zapadka, Christopher~M Lack, Jonathan~D Clemente, Christopher~T Whitlow, and Ge Wang.
\newblock Head-neck dual-energy ct contrast media reduction using diffusion models.
\newblock \emph{arXiv preprint arXiv:2308.13002}, 2023{\natexlab{b}}.

\bibitem[Montalt-Tordera et~al.(2021)Montalt-Tordera, Quail, Steeden, and Muthurangu]{montalt2021reducing}
Javier Montalt-Tordera, Michael Quail, Jennifer~A Steeden, and Vivek Muthurangu.
\newblock Reducing contrast agent dose in cardiovascular mr angiography with deep learning.
\newblock \emph{Journal of Magnetic Resonance Imaging}, 54\penalty0 (3):\penalty0 795--805, 2021.

\bibitem[{\"O}zbey et~al.(2023){\"O}zbey, Dalmaz, Dar, Bedel, {\"O}zturk, G{\"u}ng{\"o}r, and {\c{C}}ukur]{ozbey2023unsupervised}
Muzaffer {\"O}zbey, Onat Dalmaz, Salman~UH Dar, Hasan~A Bedel, {\c{S}}aban {\"O}zturk, Alper G{\"u}ng{\"o}r, and Tolga {\c{C}}ukur.
\newblock Unsupervised medical image translation with adversarial diffusion models.
\newblock \emph{IEEE Transactions on Medical Imaging}, 2023.

\bibitem[{\"O}zt{\"u}rk et~al.(2024){\"O}zt{\"u}rk, Duran, and {\c{C}}ukur]{ozturk2024denomamba}
{\c{S}}aban {\"O}zt{\"u}rk, O{\u{g}}uz~Can Duran, and Tolga {\c{C}}ukur.
\newblock Denomamba: A fused state-space model for low-dose ct denoising.
\newblock \emph{arXiv preprint arXiv:2409.13094}, 2024.

\bibitem[Pesteie et~al.(2019)Pesteie, Abolmaesumi, and Rohling]{pesteie2019adaptive}
Mehran Pesteie, Purang Abolmaesumi, and Robert~N Rohling.
\newblock Adaptive augmentation of medical data using independently conditional variational auto-encoders.
\newblock \emph{IEEE transactions on medical imaging}, 38\penalty0 (12):\penalty0 2807--2820, 2019.

\bibitem[Ruan and Xiang(2024)]{ruan2024vm}
Jiacheng Ruan and Suncheng Xiang.
\newblock Vm-unet: Vision mamba unet for medical image segmentation.
\newblock \emph{arXiv preprint arXiv:2402.02491}, 2024.

\bibitem[Smith et~al.(2022)Smith, Warrington, and Linderman]{smith2022simplified}
Jimmy~TH Smith, Andrew Warrington, and Scott~W Linderman.
\newblock Simplified state space layers for sequence modeling.
\newblock \emph{arXiv preprint arXiv:2208.04933}, 2022.

\bibitem[Song et~al.(2024)Song, Hu, Luo, Fessler, and Shen]{song2024diffusionblend}
Bowen Song, Jason Hu, Zhaoxu Luo, Jeffrey~A Fessler, and Liyue Shen.
\newblock Diffusionblend: Learning 3d image prior through position-aware diffusion score blending for 3d computed tomography reconstruction.
\newblock \emph{arXiv preprint arXiv:2406.10211}, 2024.

\bibitem[Tan et~al.(2021)Tan, Zhou, Li, Yang, Chen, and Yang]{tan2021automated}
Wenjun Tan, Luyu Zhou, Xiaoshuo Li, Xiaoyu Yang, Yufei Chen, and Jinzhu Yang.
\newblock Automated vessel segmentation in lung ct and cta images via deep neural networks.
\newblock \emph{Journal of X-ray science and technology}, 29\penalty0 (6):\penalty0 1123--1137, 2021.

\bibitem[Tsai et~al.(2024)Tsai, Lin, Hu, Zhu, Wang, et~al.]{tsai2024uu}
Ting~Yu Tsai, Li Lin, Shu Hu, Hongtu Zhu, Xin Wang, et~al.
\newblock Uu-mamba: Uncertainty-aware u-mamba for cardiac image segmentation.
\newblock \emph{arXiv preprint arXiv:2405.17496}, 2024.

\bibitem[Update(2017)]{update2017heart}
AHA~Statistical Update.
\newblock Heart disease and stroke statistics--2017 update.
\newblock \emph{Circulation}, 135:\penalty0 e146--e603, 2017.

\bibitem[Wang et~al.(2024{\natexlab{a}})Wang, Yi, Wen, Zhu, and Xu]{wang2024cardiovascular}
Zhifeng Wang, Renjiao Yi, Xin Wen, Chenyang Zhu, and Kai Xu.
\newblock Cardiovascular medical image and analysis based on 3d vision: A comprehensive survey.
\newblock \emph{Meta-Radiology}, page 100102, 2024{\natexlab{a}}.

\bibitem[Wang et~al.(2024{\natexlab{b}})Wang, Zhang, Wang, and Zhang]{wang2024soft}
Zhenbin Wang, Lei Zhang, Lituan Wang, and Zhenwei Zhang.
\newblock Soft masked mamba diffusion model for ct to mri conversion.
\newblock \emph{arXiv preprint arXiv:2406.15910}, 2024{\natexlab{b}}.

\bibitem[Wang et~al.(2024{\natexlab{c}})Wang, Zheng, Zhang, Cui, and Li]{wang2024mamba}
Ziyang Wang, Jian-Qing Zheng, Yichi Zhang, Ge Cui, and Lei Li.
\newblock Mamba-unet: Unet-like pure visual mamba for medical image segmentation.
\newblock \emph{arXiv preprint arXiv:2402.05079}, 2024{\natexlab{c}}.

\bibitem[Weisbord et~al.(2020)Weisbord, Palevsky, Kaufman, Wu, Androsenko, Ferguson, Parikh, Bhatt, and Gallagher]{weisbord2020contrast}
Steven~D Weisbord, Paul~M Palevsky, James~S Kaufman, Hongsheng Wu, Maria Androsenko, Ryan~E Ferguson, Chirag~R Parikh, Deepak~L Bhatt, and Martin Gallagher.
\newblock Contrast-associated acute kidney injury and serious adverse outcomes following angiography.
\newblock \emph{Journal of the American College of Cardiology}, 75\penalty0 (11):\penalty0 1311--1320, 2020.

\bibitem[Wen et~al.(2024)Wen, Zhu, Yi, Wang, Zhu, and Xu]{wen2024cad}
Xin Wen, Xuening Zhu, Renjiao Yi, Zhifeng Wang, Chenyang Zhu, and Kai Xu.
\newblock Cad-nerf: Learning nerfs from uncalibrated few-view images by cad model retrieval.
\newblock \emph{arXiv preprint arXiv:2411.02979}, 2024.

\bibitem[Xia et~al.(2023)Xia, Ravikumar, Lassila, and Frangi]{xia2023virtual}
Yan Xia, Nishant Ravikumar, Toni Lassila, and Alejandro~F Frangi.
\newblock Virtual high-resolution mr angiography from non-angiographic multi-contrast mris: synthetic vascular model populations for in-silico trials.
\newblock \emph{Medical Image Analysis}, 87:\penalty0 102814, 2023.

\bibitem[Yang et~al.(2023)Yang, Musio, Ma, Juchler, Paetzold, Al-Maskari, H{\"o}her, Li, Hamamci, Sekuboyina, et~al.]{yang2023benchmarking}
Kaiyuan Yang, Fabio Musio, Yihui Ma, Norman Juchler, Johannes~C Paetzold, Rami Al-Maskari, Luciano H{\"o}her, Hongwei~Bran Li, Ibrahim~Ethem Hamamci, Anjany Sekuboyina, et~al.
\newblock Benchmarking the cow with the topcow challenge: Topology-aware anatomical segmentation of the circle of willis for cta and mra.
\newblock \emph{ArXiv}, 2023.

\bibitem[Yi et~al.(2019)Yi, Walia, and Babyn]{yi2019generative}
Xin Yi, Ekta Walia, and Paul Babyn.
\newblock Generative adversarial network in medical imaging: A review.
\newblock \emph{Medical image analysis}, 58:\penalty0 101552, 2019.

\bibitem[Yue and Li(2024)]{yue2024medmamba}
Yubiao Yue and Zhenzhang Li.
\newblock Medmamba: Vision mamba for medical image classification.
\newblock \emph{arXiv preprint arXiv:2403.03849}, 2024.

\bibitem[Zanzonico et~al.(2006)Zanzonico, Rothenberg, and Strauss]{zanzonico2006radiation}
Pat Zanzonico, Lawrence~N Rothenberg, and H~William Strauss.
\newblock Radiation exposure of computed tomography and direct intracoronary angiography: risk has its reward.
\newblock \emph{Journal of the American College of Cardiology}, 47\penalty0 (9):\penalty0 1846--1849, 2006.

\bibitem[Zhang et~al.(2019)Zhang, Goodfellow, Metaxas, and Odena]{zhang2019self}
Han Zhang, Ian Goodfellow, Dimitris Metaxas, and Augustus Odena.
\newblock Self-attention generative adversarial networks.
\newblock In \emph{International conference on machine learning}, pages 7354--7363. PMLR, 2019.

\bibitem[Zhang et~al.(2023)Zhang, Xu, Usuyama, Xu, Bagga, Tinn, Preston, Rao, Wei, Valluri, et~al.]{zhang2023biomedclip}
Sheng Zhang, Yanbo Xu, Naoto Usuyama, Hanwen Xu, Jaspreet Bagga, Robert Tinn, Sam Preston, Rajesh Rao, Mu Wei, Naveen Valluri, et~al.
\newblock Biomedclip: a multimodal biomedical foundation model pretrained from fifteen million scientific image-text pairs.
\newblock \emph{arXiv preprint arXiv:2303.00915}, 2023.

\bibitem[Zheng et~al.(2024)Zheng, Mo, Sun, Li, Wu, Wang, Vincent, and Papie{\.z}]{zheng2024deformation}
Jian-Qing Zheng, Yuanhan Mo, Yang Sun, Jiahua Li, Fuping Wu, Ziyang Wang, Tonia Vincent, and Bart{\l}omiej~W Papie{\.z}.
\newblock Deformation-recovery diffusion model (drdm): Instance deformation for image manipulation and synthesis.
\newblock \emph{arXiv preprint arXiv:2407.07295}, 2024.

\bibitem[Zhou et~al.(2020)Zhou, Fu, Chen, Shen, and Shao]{zhou2020hi}
Tao Zhou, Huazhu Fu, Geng Chen, Jianbing Shen, and Ling Shao.
\newblock Hi-net: hybrid-fusion network for multi-modal mr image synthesis.
\newblock \emph{IEEE transactions on medical imaging}, 39\penalty0 (9):\penalty0 2772--2781, 2020.

\bibitem[Zhu et~al.(2024)Zhu, Liao, Zhang, Wang, Liu, and Wang]{zhu2024vision}
Lianghui Zhu, Bencheng Liao, Qian Zhang, Xinlong Wang, Wenyu Liu, and Xinggang Wang.
\newblock Vision mamba: Efficient visual representation learning with bidirectional state space model.
\newblock \emph{arXiv preprint arXiv:2401.09417}, 2024.

\bibitem[Zhuo and Shen(2024)]{zhuo2024diffusereg}
Yongtai Zhuo and Yiqing Shen.
\newblock Diffusereg: Denoising diffusion model for obtaining deformation fields in unsupervised deformable image registration.
\newblock In \emph{International Conference on Medical Image Computing and Computer-Assisted Intervention}, pages 597--607. Springer, 2024.

\end{thebibliography}
}


\end{document}